\pdfoutput=1

\documentclass[11pt]{article}

\usepackage[final]{ACL2023}

\usepackage{times}
\usepackage{latexsym}

\usepackage[T1]{fontenc}

\usepackage[utf8]{inputenc}

\usepackage{microtype}
\usepackage{inconsolata}

\usepackage{url}
\usepackage{nicefrac}
\usepackage{calc}
\usepackage{xcolor}
\usepackage{graphicx}
\usepackage{multirow}
\usepackage{comment}
\usepackage{amsthm}
\usepackage{amssymb}
\usepackage{amsmath}
\usepackage{amsfonts}
\usepackage{color, colortbl}
\usepackage{adjustbox}
\usepackage{tabularx}
\usepackage{booktabs}
\usepackage{makecell}
\usepackage{xspace}
\usepackage{multirow}
\usepackage{arydshln}
\usepackage{tikz}
\usepackage{enumitem}
\usepackage{euflag}
\usepackage{subfig}
\usetikzlibrary{positioning}

\title{Cross-Lingual Transfer with Target Language-Ready Task Adapters}

\author{Marinela Parovi\'{c}$^1$ ~~~ Alan Ansell$^{1}$ ~~~ Ivan Vuli\'{c}$^{1}$ ~~~ Anna Korhonen$^{1}$\smallskip \\
$^1$Language Technology Lab, TAL, University of Cambridge \\
\texttt {\{mp939,aja63,iv250,alk23\}@cam.ac.uk}
}

\definecolor{Gray}{gray}{0.92}
\definecolor{StrongerGray}{gray}{0.84}
\newcolumntype{Y}{>{\centering\arraybackslash}X}

\newcommand{\sparagraph}[1]{\noindent\textbf{#1.}}
\newcommand{\rparagraph}[1]{\vspace{1.8mm}\noindent\textbf{#1.}}
\newcommand{\rparagraphnodot}[1]{\vspace{1.8mm}\noindent\textbf{#1}}

\usepackage{todonotes}
\makeatletter
\newcommand*\iftodonotes{\if@todonotes@disabled\expandafter\@secondoftwo\else\expandafter\@firstoftwo\fi}
\makeatother

\newcommand{\eu}{{\textsc{eu}}\xspace}
\newcommand{\myv}{{\textsc{myv}}\xspace}
\newcommand{\kpv}{{\textsc{kpv}}\xspace}
\newcommand{\mr}{{\textsc{mr}}\xspace}
\newcommand{\wo}{{\textsc{wo}}\xspace}
\newcommand{\af}{{\textsc{af}}\xspace}
\newcommand{\bm}{{\textsc{bm}}\xspace}
\newcommand{\mt}{{\textsc{mt}}\xspace}
\newcommand{\te}{{\textsc{te}}\xspace}
\newcommand{\ug}{{\textsc{ug}}\xspace}

\newcommand{\ay}{{\textsc{aym}}\xspace}
\newcommand{\gn}{{\textsc{gn}}\xspace}
\newcommand{\nah}{{\textsc{nah}}\xspace}
\newcommand{\qu}{{\textsc{quy}}\xspace}
\newcommand{\bzd}{{\textsc{bzd}}\xspace}
\newcommand{\cni}{{\textsc{cni}}\xspace}
\newcommand{\hch}{{\textsc{hch}}\xspace}
\newcommand{\oto}{{\textsc{oto}}\xspace}
\newcommand{\shp}{{\textsc{shp}}\xspace}
\newcommand{\tar}{{\textsc{tar}}\xspace}

\newcommand{\hau}{{\textsc{hau}}\xspace}
\newcommand{\ibo}{{\textsc{ibo}}\xspace}
\newcommand{\kin}{{\textsc{kin}}\xspace}
\newcommand{\lug}{{\textsc{lug}}\xspace}
\newcommand{\luo}{{\textsc{luo}}\xspace}
\newcommand{\pcm}{{\textsc{pcm}}\xspace}
\newcommand{\swa}{{\textsc{swa}}\xspace}
\newcommand{\wol}{{\textsc{wol}}\xspace}
\newcommand{\yor}{{\textsc{yor}}\xspace}

\newcommand{\ar}{{\textsc{ar}}\xspace}
\newcommand{\hi}{{\textsc{hi}}\xspace}
\newcommand{\sw}{{\textsc{sw}}\xspace}
\newcommand{\thai}{{\textsc{th}}\xspace}
\newcommand{\ur}{{\textsc{ur}}\xspace}
\newcommand{\zh}{{\textsc{zh}}\xspace}

\newcommand{\bn}{{\textsc{bn}}\xspace}

\newcommand{\badx}{{\textsc{bad-x}}\xspace}
\newcommand{\madx}{{\textsc{mad-x}}\xspace}

\newcommand{\allmulti}{{\textsc{All-Multi}}\xspace}

\begin{document}
\maketitle
\begin{abstract}
Adapters have emerged as a modular and parameter-efficient approach to (zero-shot) cross-lingual transfer. The established MAD-X framework employs separate language and task adapters which can be arbitrarily combined to perform the transfer of any task to any target language. Subsequently, BAD-X, an extension of the MAD-X framework, achieves improved transfer at the cost of MAD-X's modularity by creating `bilingual' adapters specific to the source-target language pair. In this work, we aim to take the best of both worlds by (i) fine-tuning \textit{task} adapters adapted to the target language(s) (so-called \textit{`target language-ready' (TLR)} adapters) to maintain high transfer performance, but (ii) without sacrificing the highly modular design of MAD-X. The main idea of `target language-ready' adapters is to resolve the training-vs-inference discrepancy of MAD-X: the task adapter `sees' the target language adapter for the very first time during inference, and thus might not be fully compatible with it. We address this mismatch by exposing the task adapter to the target language adapter during training, and empirically validate several variants of the idea: in the simplest form, we alternate between using the source and target language adapters during task adapter training, which can be generalized to cycling over any set of language adapters. We evaluate different TLR-based transfer configurations with varying degrees of generality across a suite of standard cross-lingual benchmarks, and find that the most general (and thus most modular) configuration consistently outperforms MAD-X and BAD-X on most tasks and languages.







\end{abstract}

\section{Introduction and Motivation}
\label{s:intro}
Recent progress in multilingual NLP has mainly been driven by massively multilingual Transformer models (MMTs) such as mBERT \citep{devlin-etal-2019-bert}, XLM-R \citep{conneau-etal-2020-unsupervised}, and mT5 \citep{xue-etal-2021-mt5}, which have been trained on the unlabeled data of 100+ languages. Their shared multilingual representation spaces enable zero-shot cross-lingual transfer \cite{pires-etal-2019-multilingual,K:2020}, that is, performing tasks with a reasonable degree of accuracy in languages that entirely lack training data for those tasks. 

Zero-shot cross-lingual transfer is typically performed by fine-tuning the pretrained MMT on task-specific data in a high-resource \textit{source} language (i.e., typically English), and then applying it directly to make task predictions in the \textit{target} language. In the standard setup, the model's knowledge about the target language is acquired solely during the pretraining stage \cite{artetxe-etal-2020-cross}. In order to improve the transfer performance, task fine-tuning can be preceded with fine-tuning on unlabeled data in the target language \cite{ponti-etal-2020-xcopa,pfeiffer-etal-2020-mad}. Nonetheless, the performance on the target languages in such scenarios is lower than that on the source language, and the difference is known as the \textit{cross-lingual transfer gap} \citep{pmlr-v119-hu20b}. Crucially, the transfer gap tends to increase for the languages where such transfer is needed the most \cite{joshi-etal-2020-state}: i.e., for low-resource target languages, and languages typologically more distant from the source language (e.g., English) \cite{lauscher-etal-2020-zero}.

\textit{Adapters} \citep{rebuffi-NIPS2017-adapters,pmlr-v97-houlsby19a} have emerged as a prominent approach for aiding zero-shot cross-lingual transfer \cite{pfeiffer-etal-2020-mad,ustun-etal-2022-udapter,ansell-etal-2021-mad-g,parovic-etal-2022-bad}. They offer several benefits: (i) providing additional representation capacity for target languages; (ii) much more parameter-efficient fine-tuning compared to full-model fine-tuning, as they allow the large MMT's parameters to remain unmodified, and thus preserve the multilingual knowledge the MMT has acquired during pretraining. They also (iii) provide modularity in learning and storing different facets of knowledge \cite{pfeiffer-etal-2020-adapterhub}: this property enables them to be combined in favorable ways to achieve better performance, and previously fine-tuned modules (e.g., language adapters) to be reused across different applications.

The established adapter-based cross-lingual transfer framework \madx \cite{pfeiffer-etal-2020-mad} trains separate language adapters (LAs) and task adapters (TAs) which can then be arbitrarily combined for the transfer of any task to any language. Despite having a highly modular design, stemming primarily from dedicated per-language and per-task adapters, \madx's TAs lack `adaptivity' to the target language(s) of interest: i.e., its TAs are fully\textit{ target language-agnostic}. More precisely, during task fine-tuning, the \madx TA is exposed only to the source language LA, and `sees' the target language TA and examples from that language for the first time only at inference. This deficiency might result in incompatibility between the TA and the target LA, which would emerge only at inference.  


\badx \cite{parovic-etal-2022-bad} trades off \madx's high degree of modularity by introducing \textit{`bilingual'} language adapters specialized for transfer between the source-target language pair.\footnote{Similarly, such bilingual adapters have been used in multilingual NMT research to boost translation between particular language pairs \cite{bapna-firat-2019-simple,philip-etal-2020-monolingual}.} While such transfer direction specialization results in a better performance, the decrease in modularity results in much larger computational requirements: \badx requires fine-tuning a dedicated bilingual LA for every language pair of interest followed up by fine-tuning a dedicated TA again for each pair.

Prior work has not explored whether this specialization (i.e., exposing the target language at training time) can be done successfully solely at the level of TAs whilst preserving modularity at the LA level. Such specialization in the most straightforward bilingual setup still requires fine-tuning a dedicated TA for each target language of interest. However, this is already a more pragmatic setup than \badx since TAs are much less computationally expensive to train than LAs. Moreover, as we show in this work, it is possible to also extend TA fine-tuning to more target languages, moving from bilingual specialization to the more universal multilingual `exposure' and towards \textit{multilingual language-universal TAs}. 

In this work, we aim to create a modular design inspired by \madx while seeking to reap the benefits of the exposure to one or more target languages. To this end, we thus introduce \textit{target language-ready (TLR)} task adapters designed to excel at a particular target language or at a larger set of target languages. In the simplest bilingual variant, TLR TAs are trained by alternating between source and target LAs, while the more general version allows cycling over any set of LAs. Creating TLR TAs does not require any expensive retraining or alternative training of LAs.

We run experiments with a plethora of standard benchmarks focused on zero-shot cross-lingual transfer and low-resource languages, covering 1) NER on MasakhaNER; 2) dependency parsing (DP) on Universal Dependencies; 3) natural language inference (NLI) on AmericasNLI and XNLI; 4) QA on XQuAD and TyDiQA-GoldP. Our results show that TLR TAs outperform \madx and \badx on all tasks on average, and offer consistent gains across a large majority of the individual target languages. Importantly, the most general TLR TA, which is shared between all target languages and thus positively impacts modularity and reusability, shows the strongest performance across the majority of tasks and target languages. Fine-tuning the TA in such multilingual setups also acts as a \textit{multilingual regularization} \cite{ansell-etal-2021-mad-g}:  while the TA gets exposed to different target languages (i.e., maintaining its TLR property), at the same time it does not overfit to a single target language as it is forced to adapt to more languages, and thus learns more universal cross-language features. Our code and models are publicly available at: \url{https://github.com/parovicm/tlr-adapters}.


\section{Methodology}
\label{s:methodology}
\subsection{Background}
\label{ss:background}
\sparagraph{Adapters} 
Following \madx and \badx, in this work we focus on the most common adapter architecture, \textit{serial adapters} \cite{pmlr-v97-houlsby19a,pfeiffer-etal-2021-adapterfusion}, but we remind the reader that other adapter options are available \cite{He:2022iclr} and might be used in the context of cross-lingual transfer. Serial adapters are lightweight bottleneck modules inserted within each Transformer layer. The architecture of an adapter at each layer consists of a down-projection, a non-linearity and an up-projection followed by a residual connection. Let the down-projection at layer $l$ be a matrix $\mathbf{D}_l \in \mathbb{R}^{h \times d}$ and the up-projection be a matrix $\mathbf{U}_l \in \mathbb{R}^{d \times h}$ where $h$ is the hidden size of the Transformer and $d$ is the hidden size of the adapter. If we denote the hidden state and the residual at layer $l$ as $\mathbf{h}_l$ and $\mathbf{r}_l$ respectively, the adapter computation of layer $l$ is then given by:
%
\begin{equation}
 A_l(\mathbf{h}_l, \mathbf{r}_l) = \mathbf{U}_l(\text{ReLU}(\mathbf{D}_l(\mathbf{h}_l))) + \mathbf{r}_l,
\end{equation}
with ReLU as the activation function.

\rparagraph{\madx and \badx Frameworks}
\madx trains dedicated LAs and TAs \cite{pfeiffer-etal-2020-mad}. LAs are trained using unlabeled Wikipedia data with a masked language modeling (MLM) objective. TAs are trained using task-specific data in the source language. Given a source language $L_s$ and a target language $L_t$, \madx trains LAs for both $L_s$ and $L_t$. The TA is trained while stacked on top of the $L_s$ LA, which is frozen. To make predictions on $L_t$, the $L_s$ LA is swapped with the $L_t$ LA.

Unlike \madx, which is based on monolingual adapters, \badx trains bilingual LAs \cite{parovic-etal-2022-bad}. A bilingual LA is trained on the unlabeled data of both $L_s$ and $L_t$ and the TA is then trained on task-specific data in $L_s$, stacked on top of the bilingual LA. To perform inference on the task in $L_t$, the same configuration is kept since the bilingual LA `knows' both $L_s$ and $L_t$.

\subsection{Target Language-Ready Task Adapters}
\label{ss:tlr}

\begin{figure}[!t]
    \centering
    \includegraphics[width=0.63\columnwidth]{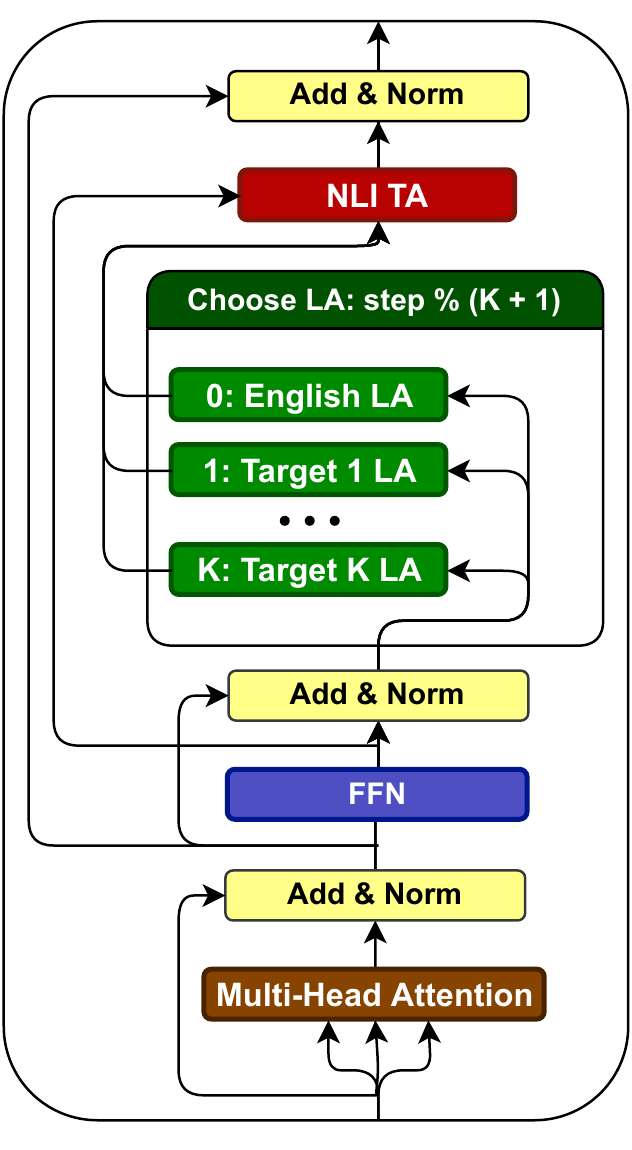}
    \caption{A general \textit{multilingual} task adapter (TA) \textit{target language-ready} (TLR) module at one MMT layer, showing the language adapters (LAs) for English as the source language and $K$ target languages along with the NLI TA. The TA is trained by cycling over the $K+1$ LAs associated with the $K+1$ languages. For a given \textit{step} number, only the LA $\textit{step}\ \%\ (K + 1)$ is switched on and the forward pass goes through that LA. Setting $K=0$ results in the original \madx setup, where only the source LA is switched on, while a bilingual TLR variant is given by $K=1$. Setting $K=1$ and removing the English LA formulates the \textsc{target}-only TLR variant. See~\S\ref{ss:tlr} for the descriptions of all the variants. The same adapter configuration(s), but with different parameters, are added at each MMT layer.}
    \label{fig:model}
\end{figure}

Instead of sacrificing the LAs' modularity as in \badx, it might be more effective to keep \madx's language-specific LAs and opt to prepare only the TAs to excel at a particular target language $L_t$, or a set of target languages of interest. Assuming LAs are available for the source language $L_s$ and $K$ target languages $L_{t,i},\ i=1,\ldots,K$, we cycle over all $K+1$ LAs during TA training, resulting in the so-called \textit{multilingual TLR TA}. This general idea is illustrated in Figure~\ref{fig:model}. The bilingual variant with a TLR TA trained by alternating between the source and target LA is a special case of the multilingual variant where $K=1$, while the original \madx setup is obtained by setting $K=0$.\footnote{It is also possible to train a TA directly without relying on any LA at all. However, previous research \cite{ansell-etal-2021-mad-g} has empirically validated that this \textit{`TA-only'} variant is consistently outperformed by \madx; hence, we do not discuss nor compare to `TA-only' in this work.}

This procedure exposes a single target language (bilingual TLR TA) or multiple target languages (multilingual TLR TA) to the TA as soon as its fine-tuning phase, making it better equipped (i.e., \textit{ready}) for the inference phase, where the TA is combined with the single $L_t$ LA.


\rparagraph{TLR Variants}
While \textsc{bilingual} TA fine-tuning follows naturally from \badx, and it seems suitable for transfer between a fixed pair of $L_s$ and $L_t$, it might be better to train the TA only on top of the $L_t$ LA. Such \textsc{target}-only TLR TAs could be particularly effective for higher-resource languages whose LAs have been trained on sufficient corpora, to the extent that pairing them with $L_s$ is detrimental. This could be especially detectable for higher-resource $L_t$-s that are also distant from $L_s$ or lack adequate vocabulary overlap with it.

\textsc{target} and \textsc{bilingual} TLR TAs require training of dedicated TAs for every $L_t$ of interest, which makes them computationally less efficient than \madx, and they introduce more parameters overall. Using \textsc{multilingual} TLR TAs mitigates this overhead. We consider two variants of \textsc{multilingual} TAs. First, the so-called \textsc{task-multi} TLR variant operates over the source language and the set of all target languages available for the task under consideration (e.g., all languages represented in the MasakhaNER dataset). Second, the \textsc{all-multi} TLR variant combines the source language with all target languages across datasets of multiple tasks (e.g., all languages represented in MasakhaNER, all languages represented in AmericasNLI, etc.); see \S\ref{s:experiments} later. These variants increase modularity and parameter efficiency and are as modular and parameter-efficient as \madx per each task: a single TA is required to handle transfer to any target language. At the same time, unlike \madx, they are offered some exposure to the representations arising from the multiple target languages they will be used for. Handling multiple LAs at fine-tuning might make the TAs more robust overall: multilinguality might act as a regularization forcing the TA to focus on more universal cross-language features~\cite{ansell-etal-2021-mad-g}.



\section{Experimental Setup}
\label{s:experiments}
\sparagraph{Evaluation Tasks and Languages} 
We comprehensively evaluate our TLR adapter framework on a suite of standard cross-lingual transfer benchmarks. They span four different task families (NER, DP, NLI and QA), with a total of six different datasets and 35 different target languages, covering a typologically and geographically diverse language sample of both low- and high-resource languages.

For NER, we use the MasakhaNER dataset \citep{adelani-etal-2021-masakhaner} which contains 10 low-resource languages from the African continent.\footnote{We exclude Amharic from our experiments as it uses a script not supported by mBERT, resulting in 9 NER target languages.} For DP, we use Universal Dependencies 2.7 \citep{zeman-2020-ud} and inherit the set of 10 typologically diverse low-resource target languages from \badx \citep{parovic-etal-2022-bad}. For NLI, we rely on the AmericasNLI dataset \citep{ebrahimi-etal-2022-americasnli}, containing 10 low-resource languages from the Americas, as well as a subset of languages from XNLI \citep{conneau-etal-2018-xnli}. Finally, for QA we use subsets of languages from XQuAD \citep{artetxe-etal-2020-cross} and TyDiQA-GoldP \citep{clark-etal-2020-tydi}. The subsets for XNLI, XQuAD and TyDiQA-GoldP were selected to combine (i) low-resource languages \cite{joshi-etal-2020-state}, with (ii) higher-resource languages for which dedicated (i.e., `\madx') LAs were readily available. The full overview of all tasks, datasets, and languages with their language codes is provided in Table~\ref{tab:tasks} in Appendix~\ref{app:tasks-and-langs}.


\rparagraph{Underlying MMT} 
We report results on all tasks with mBERT, pretrained on Wikipedias of 104 languages~\citep{devlin-etal-2019-bert}.  mBERT has been suggested by prior work as a better-performing MMT for truly low-resource languages \cite{pfeiffer-etal-2021-unks,ansell-etal-2021-mad-g}. To validate the robustness of our TLR adapters, we also use XLM-R \citep{conneau-etal-2020-unsupervised} for a subset of tasks.




\rparagraph{Language Adapters}
We train LAs for the minimum of 100 epochs or 100,000 steps with a batch size of 8, a learning rate of $5 \cdot 10^{-5}$ and a maximum sequence length of 256.\footnote{For some low-resource languages with small corpora 100 epochs leads to under-training, so the minimum number of training steps is set to 30,000.} We evaluate the LAs every 1,000 steps for low-resource languages and every 5,000 steps for high-resource ones, and choose the LA that yields the lowest perplexity, evaluated on the 5\% of the held-out monolingual data (1\% for high-resource languages). For the \badx baseline, we directly use the bilingual LAs from \citep{parovic-etal-2022-bad}.  Following \citet{pfeiffer-etal-2020-mad}, the adapter reduction factor (i.e., the ratio between MMT's hidden size and the adapter's bottleneck size) is 2 for all LAs. For the \madx LAs, we use the efficient Pfeiffer adapter configuration \cite{pfeiffer-etal-2020-adapterhub} with invertible adapters, whereas \badx LAs do not include them.


\setlength{\tabcolsep}{5pt}
\begin{table}[!t]
\def\arraystretch{1.05}
\centering
{\small
\resizebox{1\linewidth}{!}{%
\begin{tabular}{l cccc}
\toprule
\textbf{} & {\bf NER} & {\bf DP} & {\bf NLI} & {\bf QA} \\

\cmidrule(lr){2-5}
{Batch Size} & {8} & {8} & {32} & {16} \\
{Epochs} & {10} & {10} & {5} & {15} \\
{Learning Rate} & {$5 \cdot 10^{-5}$} & {$5 \cdot 10^{-5}$} & {$2 \cdot 10^{-5}$} & {$10^{-4}$} \\
{Eval Freq. (steps)} & {250} & {250} & {625} & {625} \\
{Eval Metric} & {F1} & {LAS} & {Acc} & {F1} \\
\bottomrule
\end{tabular}
}%
}%
\vspace{-0.5mm}
\caption{Hyperparameters for different tasks.}
\label{tab:task-hyperparams}
\vspace{-1mm}
\end{table}

\rparagraph{Task Adapters} 
We fine-tune TAs by stacking them on top of the corresponding LAs (see Figure~\ref{fig:model}). During their fine-tuning, the MMT's parameters and all the LAs' parameters are frozen. The adapter reduction factor for all TAs is 16 as in prior work \cite{pfeiffer-etal-2020-mad} (i.e., $d=48$), and, like the LAs, they use the Pfeiffer configuration. The hyperparameters across different tasks, also borrowed from prior work, are listed in Table~\ref{tab:task-hyperparams}. In addition, we use early stopping of 4 when training the QA TA (i.e., we stop training when the F1 score does not increase for the four consecutive evaluation cycles). We use the English SQuADv1.1 training data \cite{rajpurkar-etal-2016-squad} for TyDiQA-GoldP since (i) it is much larger than TyDiQA's native training set, and (ii) we observed higher performance on target languages in our preliminary experiments than with TyDiQA's training data.

\rparagraph{Transfer Setup: Details}
In all our transfer experiments, the source language $L_s$ is fixed to English, and we evaluate different variants described in \S\ref{ss:tlr}. For the \madx baseline, we rely on its `\madx v2.0' variant, which drops the adapters in the last layer of the Transformer, which has been found to improve transfer performance across the board \cite{pfeiffer-etal-2021-unks}. For the \textsc{task-multi} TLR variant, along with using the English LA, we fine-tune TAs using the LAs of all our evaluation languages in that particular dataset. For instance, for DP this spans 10 languages, while for NLI, we fine-tune a separate \textsc{task-multi} TLR with the 10 languages from AmericasNLI, and another one for the XNLI languages. For the \textsc{all-multi} TLR variant, in addition to English LA, we cycle over the LAs of all our evaluation languages from all the tasks and datasets.


\section{Results and Discussion}
\label{s:discussion}
\begin{figure}[!t]
    \centering
    \includegraphics[width=0.99\columnwidth]{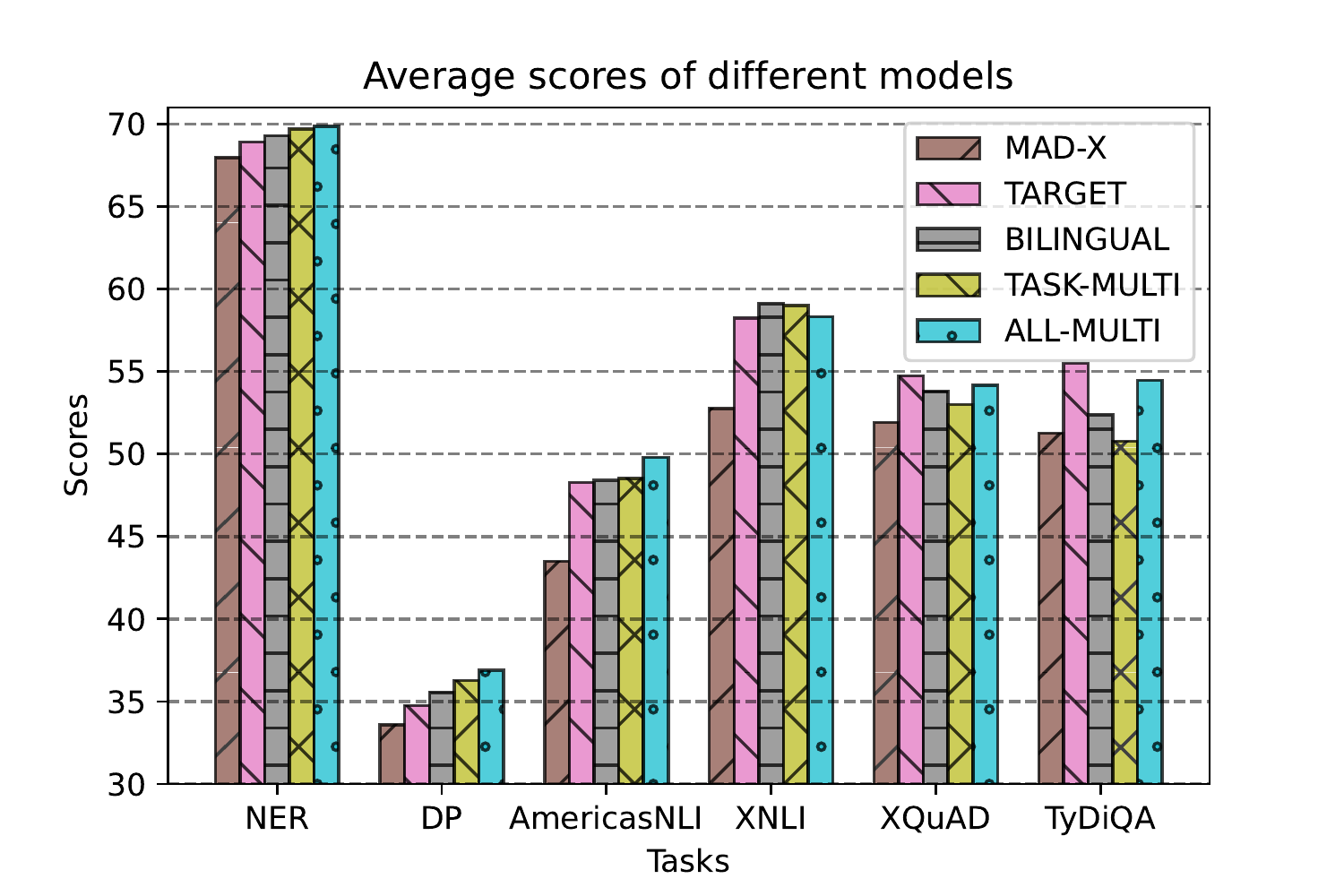}
    \vspace{-1.5mm}
    \caption{The average scores of \madx, \textsc{Target}, \textsc{Bilingual}, \textsc{Task-Multi} and \textsc{All-Multi} variants on NER (F1), DP (LAS), AmericasNLI (acc), XNLI (acc), XQuAD (F1) and TyDiQA (F1) datasets.}
    \label{fig:avg-results}
\end{figure}

\sparagraph{Main Results}
The main results with mBERT for all tasks and all languages are shown in Table~\ref{tab:results-all}, with the averages concisely provided in Figure~\ref{fig:avg-results}. Additional results with XLM-R are available in Appendix~\ref{app:xlmr}. As a general trend, we observe that all proposed TLR variants outperform \madx on the majority of the target languages across all tasks. Besides reaching higher averages on all tasks, the best per-task variants from the TLR framework surpass \madx on: 9/9 (NER), 10/10 (DP), 10/10 (AmericasNLI), 6/6 (XNLI), 4/4 (XQuAD) and 5/5 (TyDiQA) target languages. We also demonstrate that gains are achieved over the much less modular \badx on two tasks (DP, AmericasNLI) for which we had readily available \badx LAs. In sum, the comprehensive set of results from Table~\ref{tab:results-all} confirms the effectiveness and versatility of TLR adapters across a range of (typologically diverse) target languages and datasets. 

\newcommand{\nertable}{\begin{tabular}{lccccccccccc}
\toprule
\textbf{Method} & {\hau} & {\ibo} & {\kin} & {\lug} & {\luo} & {\pcm} & {\swa} & {\wol} & {\yor} & \textbf{avg} & \text{Better} \\
\cmidrule(lr){2-12}
\madx & {81.30} & {70.27} & {62.53} & {64.70} & {48.20} & {72.94} & {74.20} & {65.56} & {71.95} & {67.96} & {} \\
\textsc{Target} & {77.58} & \textbf{73.99} & {64.34} & {68.08} & {51.20} & {74.00} & {75.26} & {63.04} & {72.76} & {68.92} & {7/9} \\
\textsc{Bilingual} & {79.93} & {71.90} & {64.74} & \textbf{68.68} & {51.18} & {74.82} & {75.68} & {63.68} & \textbf{73.00} & {69.29} & {7/9} \\
\textsc{Task-Multi} & {81.83} & {72.76} & {65.03} & {66.95} & {50.69} & {75.35} & \textbf{76.59} & {65.87} & {72.26} & {69.70} & {9/9} \\
\textsc{All-Multi} & \textbf{82.39} & {71.82} & \textbf{65.12} & {66.38} & \textbf{51.38} & \textbf{76.17} & {76.42} & \textbf{66.93} & {72.10} & \textbf{69.86} & {9/9} \\
\hdashline
\textsc{Leave-Out-Task} & {82.54} & {70.88} & {65.74} & {65.78} & {49.93} & {75.33} & {76.10} & {65.27} & {72.61} & {69.35} & {8/9} \\
\textsc{Leave-Out-Targ} & {82.60} & {71.11} & {64.50} & {66.95} & {51.38} & {75.21} & {75.62} & {65.57} & {71.90} & {69.43} & {8/9} \\
\bottomrule
\end{tabular}}

\newcommand{\dptable}{\begin{tabular}{lcccccccccccc}
\toprule
\textbf{Method} & {\af} & {\bm} & {\eu} & {\kpv} & {\mr} & {\mt} & {\myv} & {\te} & {\ug} & {\wo} & \textbf{avg} & \text{Better} \\
\cmidrule(lr){2-13}
\madx & {55.21} & {13.73} & {33.20} & {23.12} & {26.18} & {47.42} & {35.70} & {49.62} & {19.60} & {32.07} & {33.59} & {} \\
\badx & {54.54} & {11.92} & {31.45} & {22.55} & {26.56} & {43.52} & {39.31} & {46.22} & {15.24} & {35.28} & {32.66} & {} \\
\textsc{Target} & {56.91} & {13.62} & {34.55} & {21.96} & {28.05} & {45.63} & {38.47} & {51.80} & {17.22} & {39.41} & {34.76} & {6/10} \\
\textsc{Bilingual} & {56.86} & {14.25} & {33.56} & {22.84} & {27.71} & {48.46} & {38.67} & {53.56} & {19.74} & {39.82} & {35.55} & {9/10} \\
\textsc{Task-Multi} & {56.56} & {15.43} & {34.90} & {22.93} & \textbf{28.70} & {51.85} & {39.18} & {53.51} & {19.48} & {40.29} & {36.28} & {8/10} \\
\textsc{All-Multi} & \textbf{57.11} & \textbf{15.46} & \textbf{35.32} & \textbf{23.76} & {28.35} & \textbf{53.68} & \textbf{39.71} & \textbf{53.83} & \textbf{20.32} & \textbf{41.34} & \textbf{36.89} & {10/10} \\
\hdashline
\textsc{Leave-Out-Task} & {56.99} & {16.40} & {33.88} & {25.27} & {28.28} & {55.03} & {39.96} & {54.11} & {21.52} & {40.41} & {37.19} & {10/10} \\
\textsc{Leave-Out-Targ} & {56.97} & {15.87} & {35.67} & {25.47} & {27.82} & {53.93} & {39.68} & {52.54} & {20.95} & {40.65} & {36.95} & {10/10} \\
\bottomrule
\end{tabular}}

\newcommand{\anlitable}{\begin{tabular}{lcccccccccccc}
\toprule
\textbf{Method} & {\ay} & {\bzd} & {\cni} & {\gn} & {\hch} & {\nah} & {\oto} & {\qu} & {\shp} & {\tar} & \textbf{avg} & \text{Better} \\
\cmidrule(lr){2-13}
\madx & {50.40} & {40.93} & {37.47} & {55.60} & {38.27} & {46.61} & {39.71} & {48.80} & {38.27} & {38.80} & {43.49} & {} \\
\badx & {46.13} & {44.67} & {45.87} & {56.80} & {44.93} & {47.70} & {41.71} & {47.87} & \textbf{49.07} & {39.47} & {46.42} & {} \\
\textsc{Target} & {50.53} & \textbf{47.20} & {44.13} & {58.00} & {43.73} & \textbf{50.54} & {41.04} & {55.87} & {46.13} & {45.47} & {48.26} & {10/10} \\
\textsc{Bilingual} & \textbf{51.73} & {46.80} & {43.07} & {58.53} & \textbf{46.13} & {48.51} & {43.32} & {55.47} & {46.00} & {44.40} & {48.40} & {10/10} \\
\textsc{Task-Multi} & {49.60} & {45.60} & {44.67} & {58.67} & {46.00} & {50.27} & {43.32} & {55.87} & {47.07} & {44.27} & {48.53} & {9/10} \\
\textsc{All-Multi} & {51.33} & \textbf{47.20} & \textbf{47.20} & \textbf{60.00} & {46.00} & {48.10} & \textbf{45.59} & \textbf{58.40} & {48.00} & \textbf{46.13} & \textbf{49.80} & {10/10} \\
\hdashline
\textsc{Leave-Out-Task} & {54.40} & {42.80} & {44.40} & {58.13} & {42.40} & {47.56} & {41.44} & {56.80} & {42.80} & {43.73} & {47.45} & {10/10} \\
\textsc{Leave-Out-Targ} & {51.07} & {44.27} & {47.33} & {59.47} & {44.53} & {47.43} & {43.98} & {56.53} & {46.53} & {42.93} & {48.41} & {10/10} \\
\bottomrule
\end{tabular}}

\newcommand{\xnlitable}{\begin{tabular}{lcccccccc}
\toprule
\textbf{Method} & {\ar} & {\hi} & {\sw} & {\thai} & {\ur} & {\zh} & \textbf{avg} & \text{Better} \\
\cmidrule(lr){2-9}
\madx & {62.75} & {56.75} & {33.33} & {43.75} & {56.41} & {63.57} & {52.76} & {} \\
\textsc{Target} & {62.87} & {57.92} & {53.93} & {52.08} & {56.79} & \textbf{65.93} & {58.25} & {6/6} \\
\textsc{Bilingual} & {63.49} & \textbf{58.62} & {54.71} & \textbf{54.95} & \textbf{57.47} & {65.49} & \textbf{59.12} & {6/6} \\
\textsc{Task-Multi} & \textbf{64.07*} & {57.88} & \textbf{55.35} & {54.19} & {56.81} & {65.69} & {59.00} & {6/6} \\
\textsc{All-Multi} & {61.98} & {57.80} & {54.15} & {53.25} & {57.05} & {65.75} & {58.33} & {5/6} \\
\bottomrule
\end{tabular}}

\newcommand{\xquadtable}{\begin{tabular}{lcccccc}
\toprule
\textbf{Method} & {\ar} & {\hi} & {\thai} & {\zh} & \textbf{avg} & \text{Better} \\
\cmidrule(lr){2-7}
\madx & {58.97/42.27} & {51.09/36.47} & {40.45/30.59} & {57.12/46.72} & {51.91/39.01} & {} \\
\textsc{Target} & {60.40/43.95} & \textbf{54.91}/40.59 & \textbf{44.95/36.22} & {58.73/48.24} & \textbf{54.75/42.25} & {4/4} \\
\textsc{Bilingual} & \textbf{60.44/44.29} & {54.18/40.42} & {42.68/33.95} & {57.95/48.32} & {53.81/41.75} & {4/4} \\
\textsc{Task-Multi} & {59.04/43.28} & {52.03/37.56} & {41.91/31.43} & \textbf{58.97}{/48.91} & {52.99/40.30} & {4/4} \\
\textsc{All-Multi} & {58.67/42.44} & {54.79/}\textbf{41.42} & {44.67/35.97} & {58.57/}\textbf{48.99} & {54.17/42.20} & {3/4} \\
\bottomrule
\end{tabular}}

\newcommand{\tyditable}{\begin{tabular}{lccccccc}
\toprule
\textbf{Method} & {\ar} & {\bn} & {\sw} & {\te} & {\thai} & \textbf{avg} & \text{Better} \\
\cmidrule(lr){2-8}
\madx & {51.10/34.42} & {56.21/42.48} & {55.04/42.49} & {46.56/34.53} & {47.41/32.91} & {51.26/37.37} & {} \\
\textsc{Target} & \textbf{56.88/40.93} & \textbf{59.47/49.56} & \textbf{61.91/50.10} & \textbf{49.92}{/39.31} & {49.36/34.81} & \textbf{55.51/42.94} & {5/5} \\
\textsc{Bilingual} & {53.50/38.65} & {53.47/40.71} & {58.26/49.10} & {48.47/38.12} & {48.22/33.67} & {52.38/40.05} & {4/5} \\
\textsc{Task-Multi} & {49.33/34.42} & {50.92/39.82} & {58.34/48.70} & {49.30/}\textbf{39.76} & {45.93/33.67} & {50.76/39.27} & {2/5} \\
\textsc{All-Multi} & {55.26/39.41} & {55.17/41.59} & {60.42/49.30} & {49.35/38.86} & \textbf{52.09/39.62} & {54.46/41.76} & {4/5} \\
\bottomrule
\end{tabular}}

\setlength{\tabcolsep}{5.3pt}
\begin{table*}
\def\arraystretch{0.93}
{\small
\begin{center}
  \subfloat[][NER: F1]{\nertable}
  \qquad
  \subfloat[][DP: LAS]{\dptable}
  \qquad
  \subfloat[][AmericasNLI: accuracy]{\anlitable}
  \qquad
  \subfloat[][XNLI: accuracy]{\xnlitable}
  \qquad
  \subfloat[][XQuAD: F1/EM]{\xquadtable}
  \qquad
  \subfloat[][TyDiQA: F1/EM]{\tyditable}
\end{center}
}
\vspace{-0.5mm}
\caption{Results of all methods and TLR variants on all tasks and target languages. 
The highest task score per each language in \textbf{bold}, but excluding the two ablation subvariants of \textsc{All-Multi} placed below the dashed horizontal lines (\textsc{Leave-Out-Task} and \textsc{Leave-Out-Targ}). \textit{Better} refers to the number of target languages for which each TLR variant scores higher than \madx. {An asterisk (*) next to the best TLR variant indicates \textit{non-significant} gains over \madx, where the significance analysis has been conducted using Student's $t$-test with $p=0.05$.}}
\label{tab:results-all}
\vspace{-1mm}
\end{table*}

\rparagraph{Breakdown of Results across Tasks and TLR Variants}
On NER and DP we observe very similar trends in results. Importantly, the most modular \textsc{All-Multi} variant offers the highest performance overall: e.g., it reaches the average F1 score of 69.86\% in the NER task, while outperforming \madx by 1.9\% on average and on all 9 target languages. Pronounced gains with that variant are also indicated in the DP task. The \textsc{Target} and \textsc{Bilingual} variants also yield gains across the majority of languages, with \textsc{Bilingual} being the stronger of the two. However, their overall utility in comparison to \textsc{All-Multi} is lower, given their lower performance coupled with lower modularity. 


On AmericasNLI, all TLR variants display considerable gains over \madx, achieving ~5-6\% higher average accuracy. They outperform \madx on all 10 target languages, except the \textsc{Task-Multi} variant with only a slight drop on \ay. The best variant is once again the most modular \textsc{All-Multi} variant, which is better than the baselines and all the other variants on 6/10 target languages.

On XNLI, which involves some higher-resource languages such as \ar, \hi and \zh, all TLR variants reach higher average accuracy than \madx. The gains peak around 5-6\% on average; however, this is due mainly to \sw where \madx completely fails, achieving the accuracy of random choice. Nonetheless, the TLR variants attain better scores on all other languages as well (the only exception is \textsc{All-Multi} on \ar). Besides \sw, \thai also marks a large boost of up to 11.2\% with the \textsc{Bilingual} variant, while the other languages attain more modest gains of up to 2\%. We remark that the \textsc{Bilingual} variant now obtains the highest average accuracy: we speculate that this could be a consequence of target languages now being on the higher-resource end compared to MasakhaNER and AmericasNLI.

Our final task family, QA, proves yet again the benefits of transfer with TLR adapters. On XQuAD and TyDiQA-GoldP, the best TLR variant is now the \textsc{Target} adapter. This might be partially due to a good representation of high-resource languages such as \ar, \hi, or \zh in mBERT and its subword vocabulary. However, we observe gains with \textsc{Target} also on lower-resource languages such as \bn and \sw on TyDiQA, which might indicate that the higher complexity of the QA task is at play in comparison to tasks such as NER and NLI.

Crucially, the most modular \textsc{All-Multi} TLR variant, which trains a single TA per each task, yields very robust and strong performance across all tasks (including the two QA tasks) and both on high-resource and low-resource languages. 


\setlength{\tabcolsep}{5pt}
\begin{table}[t]
\def\arraystretch{0.85}
\centering
{\small
\begin{tabularx}{0.99\linewidth}{l YY}
\toprule
\textbf{Method} & {DP} & {AmericasNLI} \\
\cmidrule(lr){2-3}
\madx & {31.29} & {45.33} \\
\badx & {32.66} & {46.42} \\
\textsc{Target} & {35.15} & {48.24} \\
\textsc{Bilingual} & {34.41} & {48.47} \\
\textsc{Task-Multi} & {35.86} & {48.05} \\
\textsc{All-Multi} & {36.47} & {48.49} \\
\bottomrule
\end{tabularx}

}
\vspace{-0.5mm}
\caption{
Robustness of TLR adapters. Average scores on DP and AmericasNLI when \madx LAs are trained with a different configuration and training setup. Per-language scores are available in Appendix \ref{app:marinela-adapters}.
}
\label{tab:marinela-adapters}
\vspace{-0.5mm}
\end{table}

\setlength{\tabcolsep}{5pt}
\begin{table}[t]
\def\arraystretch{0.85}
\centering
{\small
\begin{tabularx}{0.99\linewidth}{l YY}
\toprule
\textbf{Method} & {NER} & {AmericasNLI} \\
\cmidrule(lr){2-3}
\madx & {68.27} & {44.66} \\
\textsc{Target} & {68.49} & {47.92} \\
\textsc{Bilingual} & {69.24} & {48.32} \\
\textsc{Task-Multi} & {69.47} & {48.55} \\
\textsc{All-Multi} & {69.10} & {49.10} \\
\hdashline
\textsc{Leave-Out-Task} & {69.37} & {47.96} \\
\textsc{Leave-Out-Targ} & {69.13} & {48.44} \\
\bottomrule
\end{tabularx}

}
\vspace{-0.5mm}
\caption{{
Gains with TLR adapters over \madx persist when scores are averages across 3 runs (i.e. 3 different random seeds). Average scores reported, while per-language scores are provided in Appendix~\ref{app:multiple-runs}. 
}
}
\label{tab:multiple-seeds}
\vspace{-0.5mm}
\end{table}


\vspace{1.4mm}
\noindent \textbf{Towards Language-Universal Task Adapters?}
Strictly speaking, if a new $(K+1)$-th target language is introduced to our proposed TLR framework, it would be necessary to train the multilingual TLR TA anew to expose it to the new target language. In practice, massively multilingual TAs could still be applied even to languages `unseen' during TA fine-tuning (e.g., in the same way as the original \madx framework does). This violates the TLR assumption, as the TA sees the target language only at inference. However, this setup might empirically validate another desirable property of our multilingual TLR framework from Figure~\ref{fig:model}: exposing the TA at fine-tuning to a multitude of languages (and their corresponding LAs) might equip the TA with improved transfer capability even to unseen languages. Put simply, the TA will not overfit to a single target language or a small set of languages as it must learn to balance across a large and diverse set of languages; see \S\ref{s:methodology}.

We thus run experiments on MasakhaNER, UD DP, and AmericasNLI with two subvariants of the most general \textsc{All-Multi} variant. First, in the \textsc{Leave-Out-Task} subvariant, we \textit{leave out} all the LAs for the languages from the corresponding task dataset when fine-tuning the TA: e.g., for AmericasNLI, that subvariant covers the LAs of all the languages in all the datasets except those appearing in AmericasNLI, so that all AmericasNLI languages are effectively `unseen' at fine-tuning. The second subvariant, termed \textsc{Leave-Out-Targ}, leaves out only one language at a time from the corresponding dataset: e.g., when evaluating on Guarani (\textsc{gn}) in AmericasNLI, the only language `unseen' by the TA at fine-tuning is \textsc{gn} as the current inference language.

The results, summarized in Tables~\ref{tab:results-all}(a)-(c), reveal that our \textsc{multilingual} TA fine-tuning indeed increases transfer capability also for the `TA-unseen' languages, and leads towards language-universal TAs. The scores with both subvariants offer substantial gains over \madx for many languages unseen during fine-tuning and in all three tasks. This confirms that (i) \madx TAs tend to overfit to the source language and thus underperform in cross-lingual transfer, and (ii) such overfitting might get mitigated through our proposed `multilingual regularization' of the TAs while keeping the same modularity benefits. Additionally, the results also confirm the versatility of the proposed TLR framework, where strong transfer gains are achieved with different sets of languages included in multilingual TA fine-tuning: e.g., the scores with the two \textsc{Leave-Out} subvariants remain strong and competitive with the full \textsc{All-Multi} variant.

\begin{figure*}[!]
    \centering
    \includegraphics[width=1.0\linewidth]{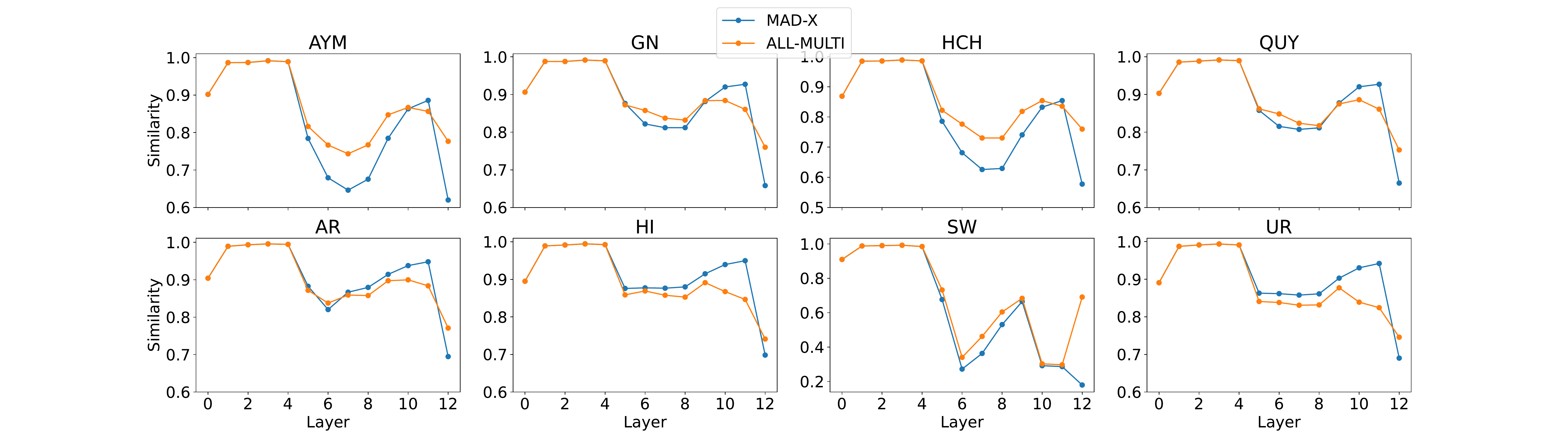}
    \caption{{Per-layer similarity scores of \madx and \allmulti adapter's representations between English and 4 languages from AmericasNLI (\ay, \gn, \hch, \qu) and 4 languages from XNLI (\ar, \hi, \sw, \ur).}}
    \label{fig:align}
\end{figure*}

For the DP task we even observe slight gains with the \textsc{Leave-Out-Task} variant over the original \textsc{All-Multi} variant which `sees' all task languages. We speculate that this might partially occur due to the phenomenon of 'the curse of multilinguality' \cite{conneau-etal-2020-unsupervised} kicking in, now at the level of the limited TA budget, but leave this for further exploration in future work.

\subsection{Further Analyses}
\sparagraph{Robustness to LA Training Configuration}
To demonstrate that our results hold even when LAs are trained with the different hyper-parameters, we adopt a training regime that makes \madx LAs directly comparable with \badx as trained in previous work by \citet{parovic-etal-2022-bad}. The average results with such LAs on DP and AmericasNLI are presented in Table~\ref{tab:marinela-adapters}, demonstrating that the gains with the proposed TLR variants hold irrespective of the LA training setup.

\rparagraph{Multiple Runs}
Given the large number of experimental runs in this work, most scores are reported from single runs with fixed seeds. However, to validate that our findings hold under different random initializations of TAs, we also run \madx and all TLR variants with three different random seeds on a subset of tasks (MasakhaNER and AmericasNLI). The main results are presented in Table~\ref{tab:marinela-adapters}, indicating that all the findings hold and are not due to a single favorable seed.

\rparagraphnodot{Do TLR Adapters Improve Alignment Between Source and Target Languages?}
In order to explain the consistent gains with TLR adapters over \madx, we analyse whether TLR adapters produce better-aligned representations between source and target languages than \madx. We execute experiments on the NLI task, choosing 4 languages from AmericasNLI (\ay, \gn, \hch, \qu) and 4 languages from XNLI (\ar, \hi, \sw, \ur) datasets, with English as a source language. 
The representations of English are obtained using MultiNLI data and English LA is paired with 1) \madx TA for the \madx baseline, and 2) \allmulti TA for the TLR representations. To obtain the representations in the target language, we use its validation data and its LA paired with either \madx TA or \allmulti TA as before. The alignment scores of both \madx and TLR methods are measured as cosine similarity between English and target representations of mBERT's $[CLS]$ token, using 500 examples in both languages. The results are presented in Figure~\ref{fig:align}. We can observe that \madx seems to have a much more significant drop in alignment values in the last layer than the \allmulti adapter, which could explain the better performance of the latter. In addition, on AmericasNLI languages, where we observe sizable gains, the \allmulti adapter seems to achieve better alignment across the middle layers of mBERT.


\section{Related Work}
\label{s:related-work}
\noindent \textbf{Parameter-Efficient Fine-Tuning} 
has emerged from an effort to overcome the need for full model fine-tuning, especially with the neural models becoming increasingly larger. Some approaches fine-tune only a subset of model parameters while keeping the rest unmodified~\citep{ben-zaken-etal-2022-bitfit,guo-etal-2021-parameter,ansell-etal-2022-composable}. Other approaches keep the model's parameters fixed and introduce a fresh set of parameters that serves for learning the desired task~\citep{li-liang-2021-prefix,lester-etal-2021-power,pmlr-v97-houlsby19a,hu2022lora}, with the tendency towards decreasing the number of newly introduced parameters while concurrently maximizing or maintaining task performance \cite{compacter,karimi-mahabadi-etal-2021-parameter}.

\vspace{1.6mm}
\noindent \textbf{Adapters} 
were introduced in computer vision research \citep{rebuffi-NIPS2017-adapters} before being brought into NLP to perform parameter-efficient transfer learning across tasks \citep{pmlr-v97-houlsby19a}. \citet{bapna-firat-2019-simple} use adapters in NMT as an efficient way of adapting the model to new languages and domains because maintaining separate models would quickly become infeasible as the number of domains and languages increases. \citet{wang-etal-2021-k} propose factual and linguistic adapters to infuse different types of knowledge into the model, while overcoming the catastrophic forgetting that would otherwise occur.

\rparagraph{Adapters for Cross-Lingual Transfer} 
\madx~\citet{pfeiffer-etal-2020-mad} introduces LAs and TAs for efficient transfer; they also propose invertible adapters for adapting MMTs to unseen languages. Subsequently, \citet{pfeiffer-etal-2021-unks} introduce a vocabulary adaptation method for \madx that can adapt the model to low-resource languages and even to unseen scripts, the latter of which was not possible with \madx's invertible adapters. In another adapter-based cross-lingual transfer approach, \citet{vidoni2020orthogonal} introduce orthogonal LAs and TAs designed to store the knowledge orthogonal to the knowledge already encoded within MMT. FAD-X \citep{lee-etal-2022-fad} explores whether the available adapters can be composed to complement or completely replace the adapters for low-resource languages. This is done through fusing \citep{pfeiffer-etal-2021-adapterfusion} TAs trained with LAs in different languages. Our TLR adapters do not involve any fusion, but rather benefit from a training procedure that operates by cycling over multiple LAs. \citet{faisal-anastasopoulos-2022-phylogeny} use linguistic and phylogenetic information to improve cross-lingual transfer by leveraging closely related languages and learning language family adapters similar to \citet{Chronopolou:2022arxiv}. This is accomplished by creating a phylogeny-informed tree hierarchy over LAs. 

UDapter \citep{ustun-etal-2020-udapter} and MAD-G \citep{ansell-etal-2021-mad-g} learn to generate LAs through the contextual parameter generation method \citep{platanios-etal-2018-contextual}. Both UDapter and MAD-G enable the generation of the parameters from vectors of typological features through sharing of linguistic information, with the main difference between the two approaches being that MAD-G's LAs are task-agnostic, while UDapter generates them jointly with a dependency parser's parameters. Hyper-X \citep{ustun2022hyper} generates weights for adapters conditioned on both task and language vectors, thus facilitating the zero-shot transfer to unseen languages and task-language combinations.

\rparagraph{Improving Cross-Lingual Transfer via Exposing Target Languages} 
In an extensive transfer case study focused on POS tagging, \citet{de-vries-etal-2022-make} showed that both source and target language (and other features such as language family, writing system, word order and lexical-phonetic distance) affect cross-lingual transfer performance. XeroAlign \citep{gritta-iacobacci-2021-xeroalign} is a method for task-specific alignment of sentence embeddings (i.e. they encourage the alignment between source task-data and its target translation by an auxiliary loss), aiming to bring the target language performance closer to that of a source language (i.e. to close the cross-lingual transfer gap). \citet{kulshreshtha-etal-2020-cross} analyze the effects of the existing methods for aligning multilingual contextualized embeddings and cross-lingual supervision, and propose a novel alignment method. \citet{yang-etal-2021-bilingual} introduce a new pretraining task to align static embeddings and multilingual contextual representations by relying on bilingual word pairs during masking. Inspired by this line of research, in this work we investigated how `exposing' target languages as well as conducting multilingual fine-tuning impacts the knowledge stored in task adapters, and their ability to boost adapter-based cross-lingual transfer.

\section{Conclusion and Future Work}
\label{s:conclusion}
We have presented a novel general framework for adapter-based cross-lingual task transfer, which improves over previous established adapter-based transfer frameworks such as \madx and \badx. The main idea is to better equip task adapters (TAs) to handle text instances in a variety of target languages. We have demonstrated that this can be achieved via so-called \textit{target language-ready} (TLR) task adapters, where we expose the TA to the target language as early as the fine-tuning stage. As another major contribution, we have also proposed a multilingual language-universal TLR TA variant which offers the best trade-off between transfer performance and modularity, learning a single universal TA that can be applied over multiple target languages. Our experiments across 6 standard cross-lingual benchmarks spanning 4 different tasks and a wide spectrum of languages have validated the considerable benefits of the proposed framework and different transfer variants emerging from it. Crucially, the most modular multilingual TLR TA variant offers the strongest performance overall, and it also generalizes well even to target languages `unseen' during TA fine-tuning. 

In future work, we plan to further investigate multilingual language-universal task adapters also in multi-task and multi-domain setups, and extend the focus from serial adapters to other adapter architectures, such as parallel adapters \cite{He:2022iclr} and sparse subnetworks \cite{ansell-etal-2022-composable,Foroutan:2022arxiv}. 



\section*{Limitations}
\label{s:limitations}

Our experiments are based on (arguably) the most standard adapter architecture for adapter-based cross-lingual transfer and beyond, which also facilitates comparisons to prior work in this area. However, we again note that there are other emerging parameter-efficient modular methods, including different adapter architectures \cite{He:2022iclr}, that could be used with the same conceptual idea. We leave further and wider explorations along this direction for future work. 


Our evaluation relies on the currently available standard multilingual benchmarks, and in particular those targeted towards low-resource languages. While the development of better models for underrepresented languages is possible mostly owing to such benchmarks, it is also inherently constrained by their quality and availability. Even though our experiments have been conducted over 35 different target languages and across several different tasks, we mostly focus on generally consistent trends across multiple languages. Delving deeper into finer-grained qualitative and linguistically oriented analyses over particular low-resource languages would require access to native speakers of those languages, and it is very challenging to conduct such analyses for many languages in our language sample.



Due to a large number of experiments across many tasks and languages, we report all our results based on a single run. Averages over multiple runs conducted on a subset of languages and tasks confirm all the core findings; for simplicity, we eventually chose to report the results for all languages and tasks in the same setup.

Finally, training language adapters is typically computationally expensive; however, owing to the modular design of our framework with respect to language adapters, these are trained only once per language and reused across different evaluations.

\section*{Acknowledgments}
We would like to thank the reviewers for their helpful suggestions.

Marinela Parović is supported by Trinity College External Research Studentship. Alan wishes to thank David and Claudia Harding for their generous support via the Harding Distinguished Postgraduate Scholarship Programme. Ivan Vuli\'{c} is supported by a personal Royal Society University Research Fellowship \textit{`Inclusive and Sustainable Language Technology for a Truly Multilingual World'} (no 221137; 2022--).

\bibliography{anthology,custom}
\bibliographystyle{acl_natbib}

\clearpage
\appendix
\label{sec:appendix}
\section{Tasks and Languages}
\label{app:tasks-and-langs}
The full list of tasks, datasets and target languages with their names and codes is given in Table \ref{tab:tasks}.

\begin{table*}[!t]
    \centering
    \def\arraystretch{0.95}
    \footnotesize
    \resizebox{\linewidth}{!}{\begin{tabular}{ m{0.15\textwidth} m{0.15\textwidth}  m{0.15\textwidth} m{0.50\textwidth} }
        \toprule
        \textbf{Task} & \textbf{Source Dataset} & \textbf{Target Dataset} & \textbf{Target Languages} \\
        \midrule
        Dependency Parsing (DP) & Universal Dependencies 2.7 \citep{zeman-2020-ud} & Universal Dependencies 2.7 \citep{zeman-2020-ud} & Afrikaans (\af)$^*$, Bambara (\bm), Basque (\eu)$^*$, Komi-Zyryan (\kpv), Marathi (\mr)$^*$, Maltese (\mt), Erzya (\myv), Telugu (\te)$^*$, Uyghur (\ug), Wolof (\wo) \\
        \midrule
        Named Entity Recognition (NER) & CoNLL 2003 \citep{tjong-kim-sang-de-meulder-2003-introduction} & MasakhaNER \citep{adelani-etal-2021-masakhaner} & Hausa (\hau), Igbo (\ibo), Kinyarwanda (\kin), Luganda (\lug), Luo (\luo), Nigerian-Pidgin (\pcm), Swahili (\swa)$^*$, Wolof (\wol), Yor\`{u}b\'{a} (\yor)$^*$ \\
        \midrule
        \multirow{2}{*}{\begin{minipage}{0.15\textwidth}Natural Language Inference (NLI)\end{minipage}}
        & MultiNLI  \citep{williams-etal-2018-broad} & AmericasNLI \citep{ebrahimi-etal-2022-americasnli} & Aymara (\ay), Bribri (\bzd), Ash\'{a}ninka (\cni), Guarani (\gn), Wixarika (\hch), N\'{a}huatl (\nah), Otom\'{i} (\oto), Quechua (\qu), Shipibo-Konibo (\shp), Rar\'{a}muri (\tar) \\
        \cmidrule(lr){2-4}
        & MultiNLI  \citep{williams-etal-2018-broad} & XNLI \citep{conneau-etal-2018-xnli} & Arabic (\ar)$^\dagger$, Hindi (\hi)$^\dagger$, Swahili (\sw)$^*$, Thai (\thai)$^\dagger$, Urdu (\ur)$^*$, Chinese (\zh)$^\dagger$ \\
        \midrule
        \multirow{2}{*}{\begin{minipage}{0.15\textwidth}Question Answering (QA)\end{minipage}}
        & SQuAD v1.1 \citep{rajpurkar-etal-2016-squad} & XQuAD  \citep{artetxe-etal-2020-cross} & Arabic (\ar)$^\dagger$, Hindi (\hi)$^\dagger$, Thai (\thai)$^\dagger$, Chinese (\zh)$^\dagger$ \\
        \cmidrule(lr){2-4}
        & SQuAD v1.1 \citep{rajpurkar-etal-2016-squad} & TyDiQA-GoldP \citep{clark-etal-2020-tydi} & Arabic (\ar)$^\dagger$, Bengali (\bn)$^*$, Swahili (\sw)$^*$, Telugu (\te)$^*$, Thai (\thai)$^\dagger$ \\
        \bottomrule
    \end{tabular}}
    \caption{Details of the tasks, datasets, and languages involved in our cross-lingual transfer evaluation. $^*$ denotes low-resource languages seen during MMT pretraining; $^\dagger$ denotes high-resource languages seen during MMT pretraining; all other languages are low-resource and unseen. The source language is always English.}
    \vspace{-1mm}
    \label{tab:tasks}
\end{table*}

\section{XLM-R Results}
\label{app:xlmr}
The results on AmericasNLI, XNLI and XQuAD with XLM-R are shown in Table \ref{tab:results-all-xlmr}.

\newcommand{\anlixlmr}{
\begin{tabular}{lcccccccccccc}
\toprule
\textbf{Method} & {\ay} & {\bzd} & {\cni} & {\gn} & {\hch} & {\nah} & {\oto} & {\qu} & {\shp} & {\tar} & \textbf{avg} & \text{Better} \\
\cmidrule(lr){2-13}
\madx & {54.40} & {40.40} & {46.80} & {58.13} & {40.80} & {48.92} & {44.39} & {55.47} & {50.67} & {42.53} & {48.25} & {} \\
\textsc{Target} & {52.67} & {43.73} & {46.13} & {58.93} & {44.80} & {49.59} & {43.45} & {57.47} & {48.67} & {41.87} & {48.73} & {5/10} \\
\textsc{Bilingual} & {53.47} & {43.47} & {47.20} & {58.40} & {44.40} & {49.73} & {41.98} & {57.73} & {47.87} & {42.27} & {48.65} & {6/10} \\
\textsc{Task-Multi} & {53.20} & {43.73} & {47.47} & {56.67} & {42.27} & {49.59} & {42.51} & {58.67} & {48.93} & {43.73} & {48.68} & {6/10} \\
\textsc{All-Multi} & {53.47} & {42.27} & {47.73} & {57.47} &{41.47} & {49.73} & {40.91} & {58.80} & {50.27} & {40.93} & {48.31} & {5/10} \\
\bottomrule
\end{tabular}}

\newcommand{\xnlixlmr}{
\begin{tabular}{lcccccccc}
\toprule
\textbf{Method} & {\ar} & {\hi} & {\sw} & {\thai} & {\ur} & {\zh} & \textbf{avg} & \text{Better} \\
\cmidrule(lr){2-9}
\madx & {66.81} & {63.89} & {64.83} & {63.41} & {60.76} & {67.43} & {64.52} & {} \\
\textsc{Target} & {67.19} & {66.37} & {63.99} & {67.05} & {61.84} & {70.40} & {66.14} & {5/6} \\
\textsc{Bilingual} & {66.67} & {66.07} & {64.37} & {66.67} & {61.68} & {70.04} & {65.92} & {4/6} \\
\textsc{Task-Multi} & {68.00} & {65.89} & {64.19} & {66.01} & {61.30} & {69.58} & {65.83} & {5/6} \\
\textsc{All-Multi} & {67.84} & {66.11} & {64.89} & {65.67} & {61.82} & {69.34} & {65.95} & {6/6} \\
\bottomrule
\end{tabular}}

\newcommand{\xquadxlmr}{
\begin{tabular}{lcccccc}
\toprule
\textbf{Method} & {\ar} & {\hi} & {\thai} & {\zh} & \textbf{avg} & \text{Better} \\
\cmidrule(lr){2-7}
\madx & {65.23/47.65} & {67.15/51.09} & {69.26/59.08} & {64.01/55.13} & {66.41/53.24} & {} \\
\textsc{Target} & {65.63/48.40} & {69.49/53.78} & {69.38/58.57} & {64.09/54.71} & {67.15/53.87} & {4/4} \\
\textsc{Bilingual} & {65.85/48.91} & {68.27/52.86} & {70.31/60.50} & {64.57/55.55} & {67.25/54.45} & {4/4} \\
\textsc{Task-Multi} & {66.23/48.40} & {68.43/52.61} & {70.25/60.42} & {65.32/56.22} & {67.56/54.41} & {4/4} \\
\textsc{All-Multi} & {65.98/49.24} & {68.24/51.60} & {67.15/56.55} & {63.07/52.94} & {66.11/52.58} & {2/4} \\
\bottomrule
\end{tabular}}

\setlength{\tabcolsep}{5.3pt}
\begin{table*}
\def\arraystretch{0.93}
{\small
\begin{center}
  \subfloat[][AmericasNLI: accuracy]{\anlixlmr}
  \qquad
  \subfloat[][XNLI: accuracy]{\xnlixlmr}
  \qquad
  \subfloat[][XQuAD: F1/EM]{\xquadxlmr}
\end{center}
}
\vspace{-0.5mm}
\caption{XLM-R: Results of all methods and TLR variants on all target languages. 
}
\label{tab:results-all-xlmr}
\vspace{-1mm}
\end{table*}

\section{\madx Adapters Trained with a Different Setup}
\label{app:marinela-adapters}
The results of \madx adapters trained in a different setup \cite{parovic-etal-2022-bad} on DP and AmericasNLI are given in Table~\ref{tab:results-all-marinela}. The results of these adapters are directly comparable with the \badx baseline, as they follow the same training setup and their summary is given in Table~\ref{tab:marinela-adapters}.

\newcommand{\dptablemarinela}{\begin{tabular}{lcccccccccccc}
\toprule
\textbf{Method} & {\af} & {\bm} & {\eu} & {\kpv} & {\mr} & {\mt} & {\myv} & {\te} & {\ug} & {\wo} & \textbf{avg} & \text{Better} \\
\cmidrule(lr){2-13}
\madx & {54.23} & {11.80} & {32.51} & {22.44} & {24.24} & {44.71} & {35.45} & {45.47} & {15.67} & {26.38} & {31.29} & {} \\
\badx & {54.54} & {11.92} & {31.45} & {22.55} & {26.56} & {43.52} & {39.31} & {46.22} & {15.24} & {35.28} & {32.66} & {} \\
\textsc{Target} & {55.07} & {11.96} & {33.31} & {20.82} & {28.05} & {48.83} & {41.75} & \textbf{52.34} & {18.60} & {40.75} & {35.15} & {9/10} \\
\textsc{Bilingual} & {54.75} & {11.86} & {33.21} & {22.09} & {26.60} & {48.74} & {38.82} & {49.86} & {16.89} & {41.27} & {34.41} & {9/10} \\
\textsc{Task-Multi} & \textbf{56.55} & {11.94} & {34.17} & {23.82} & {27.71} & {51.66} & {40.87} & {51.10} & \textbf{18.90} & {41.93} & {35.86} & {10/10} \\
\textsc{All-Multi} & {56.28} & \textbf{12.91} & \textbf{35.04} & \textbf{24.11} & \textbf{28.28} & \textbf{53.02} & \textbf{41.85} & {51.43} & {18.47} & \textbf{43.31} & \textbf{36.47} & {10/10} \\
\bottomrule
\end{tabular}}

\newcommand{\anlitablemarinela}{\begin{tabular}{lcccccccccccc}
\toprule
\textbf{Method} & {\ay} & {\bzd} & {\cni} & {\gn} & {\hch} & {\nah} & {\oto} & {\qu} & {\shp} & {\tar} & \textbf{avg} & \text{Better} \\
\cmidrule(lr){2-13}
\madx & {47.07} & \textbf{45.07} & {41.87} & {55.33} & {39.47} & {48.51} & {40.91} & {51.47} & {41.60} & {42.00} & {45.33} & {} \\
\badx & {46.13} & {44.67} & {45.87} & {56.80} & {44.93} & {47.70} & {41.71} & {47.87} & \textbf{49.07} & {39.47} & {46.42} & {} \\
\textsc{Target} & {48.80} & {44.80} & {44.13} & {58.27} & {43.73} & \textbf{51.90} & {41.84} & {57.47} & {46.40} & {45.07} & {48.24} & {9/10} \\
\textsc{Bilingual} & \textbf{49.87} & {44.13} & {45.87} & {60.40} & {43.47} & {50.27} & {41.98} & \textbf{58.00} & {46.53} & {44.13} & {48.47} & {9/10} \\
\textsc{Task-Multi} & {46.40} & {44.27} & {45.87} & {57.60} & {44.40} & {50.68} & {42.78} & \textbf{58.00} & {46.53} & {44.00} & {48.05} & {8/10} \\
\textsc{All-Multi} & {46.00} & {44.00} & \textbf{46.40} & \textbf{61.07} & \textbf{46.53} & {49.32} & \textbf{44.12} & {55.33} & {46.67} & \textbf{45.47} & \textbf{48.49} & {8/10} \\
\bottomrule
\end{tabular}}

\setlength{\tabcolsep}{5.3pt}
\begin{table*}
\def\arraystretch{0.93}
{\small
\begin{center}
  \subfloat[][DP: LAS]{\dptablemarinela}
  \qquad
  \subfloat[][AmericasNLI: accuracy]{\anlitablemarinela}
\end{center}
}
\vspace{-0.5mm}
\caption{
Results of all methods and TLR variants on DP and AmericasNLI across all target languages. 
All adapters in these experiments have been trained using the hyperparameters from \citet{parovic-etal-2022-bad}.
The highest task score per each language is in \textbf{bold}. \textit{Better} refers to the number of target languages for which each TLR variant scores higher than \madx.}

\label{tab:results-all-marinela}
\vspace{-1mm}
\end{table*}

\section{Per-Language Results with Multiple Runs}
\label{app:multiple-runs}
Full results on MasakhaNER and AmericasNLI for all target languages obtained as an average across 3 different random seeds are given in Table~\ref{tab:results-all-seeds}.

\newcommand{\nertableseeds}{\begin{tabular}{lcccccccccccc}
\toprule
\textbf{Method} & {\hau} & {\ibo} & {\kin} & {\lug} & {\luo} & {\pcm} & {\swa} & {\wol} & {\yor} & \textbf{avg} & \text{Better} \\
\cmidrule(lr){2-12}
\madx & \textbf{82.00} & {70.92} & {63.55} & {65.26} & {48.62} & {72.40} & {74.53} & {64.35} & {72.78} & {68.27} & {} \\
\textsc{Target} & {78.32} & {71.70} & {63.35} & {67.52} & \textbf{50.88} & {73.99} & {75.46} & {62.55} & {72.68} & {68.49} & {5/9} \\
\textsc{Bilingual} & {80.68} & {71.56} & {63.92} & \textbf{68.11} & {50.49} & \textbf{74.78} & \textbf{76.43} & {64.39} & \textbf{72.80} & {69.24} & {8/9} \\
\textsc{Task-Multi} & {81.85} & \textbf{72.18} & \textbf{65.39} & {66.98} & {50.61} & {74.42} & {76.14} & \textbf{65.58} & {72.07} & \textbf{69.47} & {7/9} \\
\textsc{All-Multi} & {81.49} & {71.32} & {64.86} & {66.26} & {50.68} & {74.42} & {75.70} & {65.52} & {71.66} & {69.10} & {7/9} \\
\hdashline
\textsc{Leave-Out-Task} & {82.30} & {70.79} & {65.61} & {67.50} & {50.81} & {74.24} & {75.69} & {65.32} & {72.08} & {69.37} & {7/9} \\
\textsc{Leave-Out-Targ} & {82.41} & {70.66} & {65.35} & {67.38} & {50.95} & {73.90} & {75.52} & {64.86} & {71.18} & {69.13} & {7/9} \\
\bottomrule
\end{tabular}}

\newcommand{\anlitableseeds}{\begin{tabular}{lcccccccccccc}
\toprule
\textbf{Method} & {\ay} & {\bzd} & {\cni} & {\gn} & {\hch} & {\nah} & {\oto} & {\qu} & {\shp} & {\tar} & \textbf{avg} & \text{Better} \\
\cmidrule(lr){2-13}
\madx & {51.55} & {41.24} & {39.47} & {56.62} & {40.09} & {45.98} & {40.82} & {49.29} & {40.71} & {40.84} & {44.66} & {} \\
\textsc{Target} & {50.89} & \textbf{46.62} & {43.42} & {57.20} & {43.42} & \textbf{49.37} & {41.31} & {56.31} & {46.62} & {44.00} & {47.92} & {9/10} \\
\textsc{Bilingual} & \textbf{53.69} & {46.18} & {43.60} & {58.40} & {44.31} & {47.92} & {42.96} & {56.00} & {46.98} & {43.20} & {48.32} & {10/10} \\
\textsc{Task-Multi} & {51.11} & {45.38} & {44.80} & {58.49} & {45.51} & {49.05} & {42.96} & {56.31} & {47.65} & \textbf{44.22} & {48.55} & {9/10} \\
\textsc{All-Multi} & {52.62} & {45.69} & \textbf{45.91} & \textbf{59.07} & \textbf{45.78} & {48.51} & \textbf{45.01} & \textbf{56.84} & \textbf{47.82} & {43.78} & \textbf{49.10} & {10/10} \\
\hdashline
\textsc{Leave-Out-Task} & {53.91} & {43.60} & {45.78} & {57.87} & {42.80} & {47.56}  & {42.87} & {56.40} & {46.13} & {42.66} & {47.96} & {10/10} \\
\textsc{Leave-Out-Targ} & {52.09} & {44.98} & {45.91} & {58.13} & {44.44} & {48.74} & {44.43} & {56.13} & {46.98} & {42.58} & {48.44} & {10/10} \\
\bottomrule
\end{tabular}}

\setlength{\tabcolsep}{5.3pt}
\begin{table*}
\def\arraystretch{0.93}
{\small
\begin{center}
  \subfloat[][NER: F1]{\nertableseeds}
  \qquad
  \subfloat[][AmericasNLI: accuracy]{\anlitableseeds}
\end{center}
}
\vspace{-0.5mm}
\caption{
Averages across 3 different random seeds of all methods and TLR variants on MasakhaNER and AmericasNLI across all target languages. 
The highest task score per each language is in \textbf{bold}, but excluding the two ablation subvariants of \textsc{All-Multi} placed below the dashed horizontal lines (\textsc{Leave-Out-Task} and \textsc{Leave-Out-Targ}). \textit{Better} refers to the number of target languages for which each TLR variant scores higher than \madx.}
\label{tab:results-all-seeds}
\vspace{-1mm}
\end{table*}

\end{document}